\pdfoutput=1

\documentclass[11pt]{article}

\usepackage{acl}
\usepackage{times}
\usepackage{latexsym}
\usepackage{hyperref}
\usepackage[T1]{fontenc}
\usepackage{microtype}
\usepackage{subcaption}
\usepackage{adjustbox}
\usepackage{array,booktabs,ragged2e}

\usepackage[utf8]{inputenc}

\usepackage{microtype}
\newcommand{\comment}[1]{#1}

\title{Model and Data Transfer for Cross-Lingual Sequence Labelling in Zero-Resource Settings}

\author{Iker García-Ferrero \quad Rodrigo Agerri \quad German Rigau \\
         HiTZ Basque Center for Language Technologies - Ixa NLP Group \\
         University of the Basque Country UPV/EHU\\
         \{
         iker.garciaf,
         rodrigo.agerri,
         german.rigau
         \}@ehu.eus}

\begin{document}
\maketitle
\begin{abstract}
Zero-resource cross-lingual transfer approaches aim to apply supervised models
from a source language to unlabelled target languages. In this paper we perform
an in-depth study of the two main techniques employed so far for cross-lingual
zero-resource sequence labelling, based either on data or model transfer.
Although previous research has proposed translation and annotation projection
(data-based cross-lingual transfer) as an effective technique for cross-lingual
sequence labelling, in this paper we experimentally demonstrate that high
capacity multilingual language models applied in a zero-shot (model-based
cross-lingual transfer) setting consistently outperform data-based
cross-lingual transfer approaches. A detailed analysis of our results suggests
that this might be due to important differences in language use. More
specifically, machine translation often generates a textual signal which is
different to what the models are exposed to when using gold standard data,
which affects both the fine-tuning and evaluation processes. Our results also
indicate that data-based cross-lingual transfer approaches remain a competitive
option when high-capacity multilingual language models are not available.
\end{abstract}

\begin{figure}[t]
    \centering
    \includegraphics[width=0.80\linewidth]{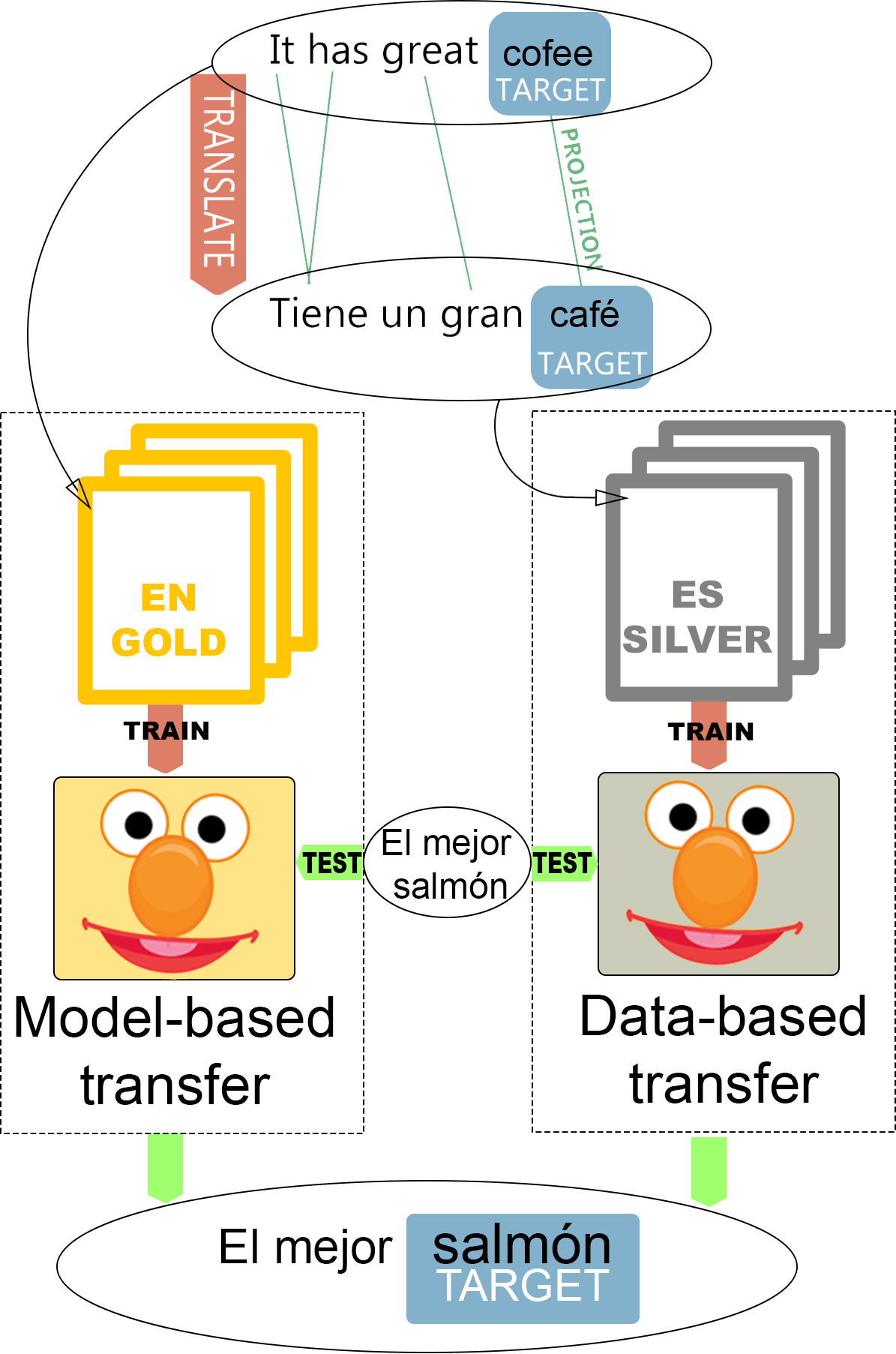}
    \caption{In the data-based transfer approach we translate and project the labels of the gold data into the target language, and use the resulting silver data to train a model for the target language. In the model-based transfer approach we train a model with gold data in English and use it in a zero-shot setting in the target language.}
    \label{fig:example}
\end{figure}

\section{Introduction}

Sequence labelling is the task of assigning a label to each token in a given input sequence. Sequence labelling is a fundamental process in many downstream NLP tasks. Currently, most successful approaches for this task apply supervised deep-neural networks \cite{lample-etal-2016-neural,akbik-etal-2018-contextual,DBLP:conf/naacl/DevlinCLT19,xlmr}. However, as it was the case for supervised statistical approaches \cite{AGERRI201663}, their performance still depends on the amount of manually annotated training data. Additionally, deep-neural models still show a significant loss of performance when evaluated in out-of-domain data \cite{DBLP:conf/aaai/Liu0YDJCMF21}. This means that to improvie their performance, it would therefore be necessary to develop very costly manually annotated data for each language and domain of application. Thus, considering that for most of the languages in the world the amount of manually annotated corpora is simply nonexistent \cite{joshi-etal-2020-state}, then the task of developing sequence labelling models for languages and domain-specific tasks, for which supervised data is not available, remains a challenge of great interest. This task is known as zero-resource cross-lingual sequence labelling.


{\bf Data-based cross-lingual transfer} methods aim to automatically generate labelled data for a target language. 
Previous works on data-based transfer have proposed \emph{translation and annotation projection} as an effective technique for zero-resource cross-lingual sequence labelling \cite{jain-etal-2019-entity,fei-etal-2020-cross}. In this setting, as illustrated in Figure \ref{fig:example}, the idea is to translate gold-labelled text into the target language and then, using automatic word alignments, project the labels from the source into the target language. The result is an automatically generated dataset in the target language that can be used for training a sequence labelling model.

The emergence of multilingual language models \cite{DBLP:conf/naacl/DevlinCLT19,xlmr} allows for {\bf model-based} cross-lingual transfer. As Figure \ref{fig:example} illustrates, using labelled data in one source language (usually English),  it is possible to fine-tune a pre-trained multilingual model that is directly used to make predictions in any of the languages included in the model. This is also known as \emph{zero-shot} cross-lingual sequence labelling. 

In this work we present an in-depth study of both approaches using the latest advancements in machine translation, word aligners and multilingual language models. We focus on two sequence labelling tasks, namely, Named Entity Recognition (NER) and Opinion Target Extraction (OTE). In order to do so, we present a data-based cross-lingual transfer approach consisting of translating gold labeled data between English and 7 other languages using state-of-the-art machine translation systems. Sequence labelling annotations are then automatically projected for every language pair. Additionally, we also produced manual alignments for those 4 languages for which we had expert annotators.
After translation and projection, for the data-transfer approach we fine-tune multilingual language models using the automatically generated datasets. We then compare the performance obtained for each of the target languages against the performance of the zero-shot cross-lingual method, consisting of fine-tuning the multilingual language models in the English gold data and generating the predictions in the required target languages.

The main contributions of our work are the following: First, we empirically establish the required conditions for each of these two approaches, data-transfer and zero-shot model-based, to outperform the other. In this sense, our experiments show that, contrary to what previous research suggested \cite{fei-etal-2020-cross,Li2021crosslingualNE}, the zero-shot model-based approach obtains the best results when high-capacity multilingual models including the target language and domain are available. Second, when the performance of the multilingual language model is not optimal for the specific target language or domain (for example when working on a text genre and domain for which available language models have not been trained), or when the required hardware to work with high-capacity language models is not easily accessible, then data-transfer based on \emph{translate and project} constitutes a competitive option. Third, we observe that machine translation data often generates training and test data which is, due to important differences in language use, markedly different to the signal received when using gold standard data in the target language. These discrepancies seem to explain the larger error rate of the translate and project method with respect to the zero-shot technique. Finally, we create manually projected datasets for four languages and automatically projected datasets for seven languages. We use them to train and evaluate cross-lingual sequence labelling models. Additionally, they are also used to extrinsically evaluate machine translation and word alignment systems. These new datasets, together with the code to generate them are publicly available to facilitate the reproducibility of results and its use in future research.\footnote{\url{https://github.com/ikergarcia1996/Easy-Label-Projection} \\ \url{https://github.com/ikergarcia1996/Easy-Translate}}

\section{Related work}
\label{sec:RelatedWord}

\subsection{Data-based cross-lingual transfer}

Data-based cross-lingual transfer methods aim to automatically generate labelled data for a target language. Some of these methods exploit parallel data. \citet{Ehrmann} automatically annotate the English version of a multi-parallel corpus and projects the annotations into all the other languages using statistical alignments of phrases. \citet{wang-manning-2014-cross} project model expectations rather than labels, which facilities transfer of model uncertainty across languages. \citet{ni-etal-2017-weakly} use a heuristic scheme that effectively selects good-quality projection-labeled data from noisy data. They also project word embeddings from a target language into a source language, so that the source-language sequence labelling system can be applied to the target language without re-training. \citet{agerri-etal-2018-building} use parallel data from multiple languages as source to project the labelled data to a target language, showing that the combination of multiple sources improves the quality of the projections. \citet{Li2021crosslingualNE} uses the XLM-R model \cite{xlmr} for labelling sequences in the source part of the parallel data and also for annotation projection. 

Instead of relying on parallel data, \citet{jain-etal-2019-entity} and \citet{fei-etal-2020-cross}, use machine translation to automatically translate the sentences of a gold-labelled dataset to the target languages. The translated data is then annotated by projecting the gold labels from the source dataset. For this purpose, \citet{jain-etal-2019-entity} first generate a list of projection candidates by orthographic and phonetic similarity. They choose the best matching candidate based on distributional statistics derived from the dataset. \citet{fei-etal-2020-cross} leverages the word alignment probabilities calculated with FastAlign \cite{fastalign} and the POS tag distributions of the source and target words. 

High quality parallel data or machine translation systems are not always available. Thus, \citet{xie-etal-2018-neural} proposes to find word translations based on bilingual word-embeddings. Alternatively, \citet{guo-roth-2021-constrained} translate labelled data in a word-by-word manner with a dictionary. Then, they the construct target-language text from the source-language annotations with a constrained pretrained language model.

\subsection{Model-based transfer}
\label{subsec:model_transfer}
Language models trained on monolingual corpora in many languages \cite{DBLP:conf/naacl/DevlinCLT19,xlmr} allow zero-shot cross-lingual model transfer. Task-specific data in one language is used to fine-tune the model for evaluation in another language \cite{pires-etal-2019-multilingual}. The zero-shot cross-lingual capability can be improved for the sequence labelling task using different techniques. 
The approaches of \citet{wang2019cross} and \citet{ouyang2021erniem} use monolingual corpora to improve the alignment of the language representations within a multilingual model.
Instead of using a single source model, \cite{rahimi-etal-2019-massively} propose to use many models from many source languages to improve the zero-shot transfer to a new language. They learn to infer which are the most reliable models in an unsupervised manner. 
\citet{wu-etal-2020-single} take advantage of a Teacher-Student learning approach. NER models in the source languages are used as teachers to train a student model on unlabeled data in the target language. 
\citet{bari2021uxla} propose an unsupervised data augmentation framework to improve the cross-lingual adaptation of models using self-training. 
\citet{hu-etal-2021-risk} use the minimum risk training framework to overcome the gap between the source and the target
languages/domains. 
They propose a unified learning algorithm based on the expectation maximization.   


Using low-capacity multilingual language models such as mBERT, \citet{fei-etal-2020-cross} finds that their data-based cross-lingual transfer approach is superior to the zero-shot transfer method. However, \citet{Li2021crosslingualNE} when using XLM-RoBERTa, a higher capacity multilingual model, obtain the best results for German and Chinese applying the data-based cross-lingual transfer approach, while the zero-shot approach is best for Spanish and Dutch. We extend their research on zero-resource settings with two different Sequence Labelling tasks, seven languages and three multilingual models of different capacity. Our experiments and the error analysis carried out establish the required conditions on which zero-shot and data-transfer approaches outperform each other. 


\comment{
}

\section{Translation and projection method}
\label{sec:Method}

\begin{figure}[t]
    \centering
    \includegraphics[width=0.99\linewidth]{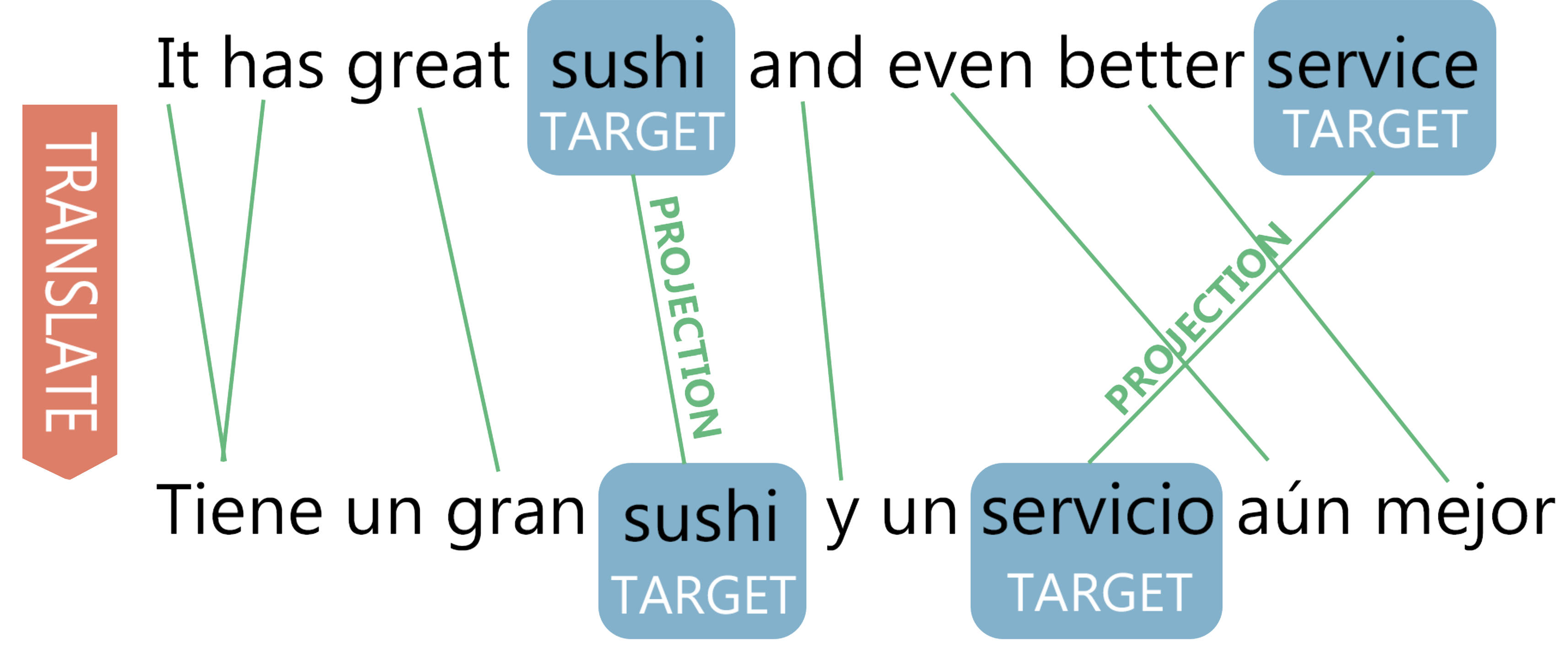}
    \caption{Illustration of the translation and annotation projection method for Opinion Target Extraction (OTE).}
    \label{fig:DataExample}
\end{figure}

Our data-based cross-lingual transfer method to perform cross-lingual sequence labelling is the following: we assume our source language to be English, for which we have train and development data. Furthermore, we also assume that the only gold-labelled data available for the target language is the evaluation set. In this setting, we automatically generate data for the target language by translating the gold-labelled English data. Then we project the gold labels from the source sentences to the translated sentences by leveraging automatic word alignments. Given a sentence  $x=\left\langle x_1,...,x_n  \right\rangle $  with length $n$ in the source language and a translated sentence    $y=\left\langle y_1,...,y_m \right\rangle$ with length $m$ in the target language,  we use a word aligner to find a set of pairs $A=\left\{\left\langle x_{i}, y_{j}\right\rangle: x_{i} \in \mathbf{x}, y_{j} \in \mathbf{y}\right\}$ where for each word pair $\left\langle x_i,y_j\right\rangle$ $y_i$ is the lexical translation of $x_j$. Next, given a sequence $s =\left\langle x_a,...,x_b  \right\rangle \in \mathbf{x}$  labeled with a category $C$ we will label the sequence $t =\left\langle y_c,...,y_d  \right\rangle \in \mathbf{y}$ with category $C$ if $\{ \forall y_j \in t \, \exists x_i \in  s : ( \left\langle x_i,y_j \right\rangle \in A )\}$. In other words, if a word labelled with a category in the source sentence is aligned to a word in the target sentence, we label the target word with the category from the word in the source sentence. Figure \ref{fig:DataExample} illustrates our method. 

When projecting annotations we find two main problems: {\it split annotations} and {\it annotation collision}. In the first case, a labeled sequence in the source sentence is split into multiple sequences in the target sentence. This happens when the alignment for a word is missing. In this case, we merge the sequences in the target sentence if the gap between them is just one word. If we still end up with multiple sequences, we choose the longest one. In the {\it annotation collision} case, a word in the target sentence is aligned to two different labelled sequences in the source language. If the two sequences are of the same category, we merge them and we label the two sequences as a single one in the target sequence. If they are of different category we just consider the one with the longest length. Finally, if a punctuation symbol in the target sequence is aligned to a labeled word in the source sentence we remove this alignment.

\section{Datasets}\label{sec:Datasets}

\begin{figure}[tbp]
  \centering
  \begin{subfigure}[Figure A]{0.49\textwidth}
   \includegraphics[width=0.99\textwidth]{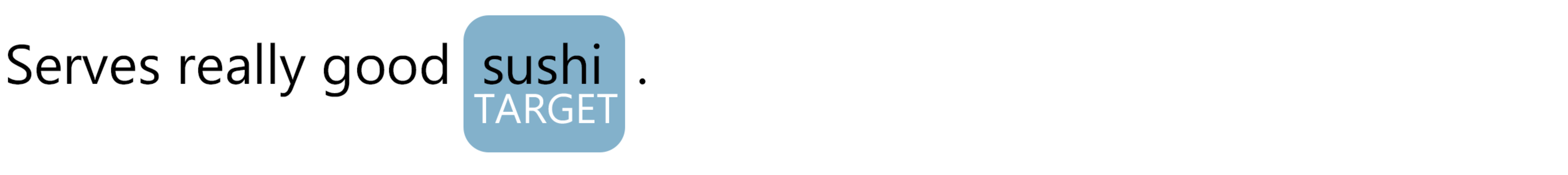}
   \caption{Illustration of the Opinion Target Extraction task.}
    \label{char:Tasks.OTE}
  \end{subfigure}%

   \begin{subfigure}[Figure B]{0.49\textwidth}
   \centering
        \includegraphics[width=0.99\textwidth]{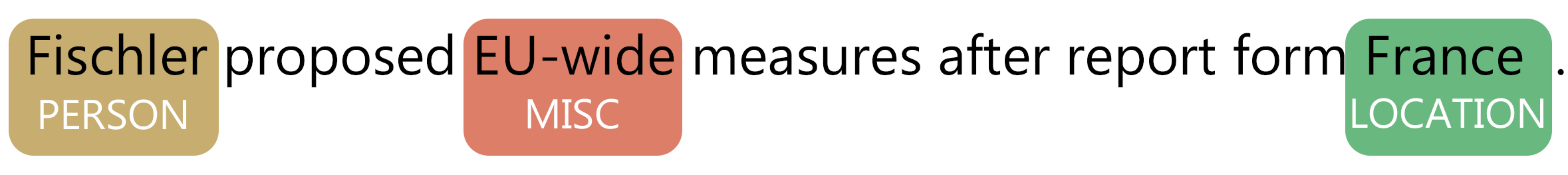}
      \caption{Illustration of the Named Entity Recognition Task.}
      \label{char:Tasks.NER}
  \end{subfigure}
  \caption{Sequence Labelling tasks used in our experiments.}
  \label{fig:Tasks}
\end{figure}

We conducted experiments in two sequence labelling tasks, namely, Opinion Target Extraction (OTE) and Named Entity Recognition (NER). Figure \ref{fig:Tasks} illustrates both tasks. 

\textbf{Opinion Target Expression (OTE)}: Given a review, the task is to detect the linguistic expression used to refer to the reviewed entity. We use the SemEval-2016 Task 5 Aspect Based Sentiment Analysis (ABSA) datasets \cite{pontiki-etal-2016-semeval}. We experiment with the English, Spanish, Dutch, French, Russian and Turkish datasets from the restaurant domain.

\textbf{Named Entity Recognition (NER)}: Given a text, the task is to detect named entities and classify them in pre-defined categories. For Spanish and Dutch we use the CoNLL-2002 datasets \cite{tjong-kim-sang-2002-introduction}. For English and German we use the CoNLL-2003 datasets \cite{tjong-kim-sang-de-meulder-2003-introduction} and for Italian we use Evalita 2009 data \cite{speranza2009named}. We map the Geo-Political Entities from Evalita 2009 to {\it location} labels to make them compatible with the CoNLL data.

\section{Experimental Setup}\label{sec:Setup}

We perform 1-to-1 annotation projection in two directions:\\
\textbf{Translate-Train}: We translate the English train and development data to the target language. We project the gold labels from the English data to the translated dataset. We then train a sequence labelling model using only the automatically generated dataset for the target language.\\
\textbf{Translate-Test}: We translate the target language test set to English. We then use a model trained in the English gold-labelled data to label the translated test set. Finally, we project the labelled sequences back to the target language.

These two data-based cross-lingual transfer approaches are compared with the \textbf{zero-shot} method in which a fine-tuned model using English gold-labelled data is evaluated by generating predictions in the target language. Finally, we also fine-tuned language models on the \textbf{gold}-labelled data, which would constitute the upper-bound in our experimental setting.

\subsection{Machine Translation}

We tested DeepL\footnote{\url{https://www.deepl.com/}}, MarianMT \cite{mariannmt,tiedemann-thottingal-2020-opus},  M2M100 (1.2B) \cite{DBLP:journals/corr/abs-2010-11125} and mBART (mbart-large-50) \cite{DBLP:journals/corr/abs-2008-00401}. A qualitative analysis performed during the projection of the OTE labels established that DeepL produced the more fluent translations. Thus, we decided to use DeepL (web version during the second half of 2021) to perform the machine translation for our data-based cross-lingual transfer experiments. The exception was Turkish, which is not supported by DeepL. In this case we use M2M100. 

\subsection{Word Alignments}

For word alignments, we use the AWESoME \cite{DBLP:journals/corr/abs-2101-08231} system. AWESoME leverages multilingual pretrained language models and fine-tune them on parallel text. Unsupervised training objectives over the parallel corpus improve the alignment quality of the models. AWESoME authors claim that the model works best with mBERT \cite{DBLP:conf/naacl/DevlinCLT19} as backbone, so we follow their advice. Although we also experimented with GIZA++ \cite{och-ney-2003-systematic}, FastAlign \cite{fastalign} and SimAlign \cite{jalili-sabet-etal-2020-simalign}, systems based on alignments from AWESoME produced the highest F1 scores when comparing the model projections and manually annotated projections (see Section \ref{sec:ErrorAnalisis}).






To train the alignment models we use the English gold-labelled dataset together with the respective MT system translations as parallel corpora. We augment the training data with 50,000 random parallel sentences from ParaCrawl v8 \cite{espla-etal-2019-paracrawl} for all the language pairs except Turkish, for which we use 50,000 random parallel sentences from the raw CCAligned v1 corpus \cite{elkishky_ccaligned_2020}.  CCAligned has received some criticism \cite{10.1162/tacl_a_00447}, but the available English-Turkish parallel data is very limited. In Section \ref{sec:ErrorAnalisis} we analyze the performance of the alignment systems, and we show that CCAligned does not hurt the performance of the aligners.

\subsection{Sequence Labelling Models}

We use three state-of-the-art multilingual pre-trained language models for sequence labelling: multilingual BERT (mBERT) \cite{DBLP:conf/naacl/DevlinCLT19} and XLM-RoBERTa (XML-R) base and large \cite{xlmr}. For both models, we add a token classification layer (linear layer) on top of each token representation. We use the sequence labelling implementation of the Huggingface open-source (Apache-2.0 License) 
library \cite{DBLP:journals/corr/abs-1910-03771}. 
F1 scores and standard deviation scores are reported by averaging the results
of 5 runs with different random seeds. Details on models sizes, hyper-parameters and datasets are provided in the Appendix
(\ref{apen:ModelSize}, \ref{sec:appendix-hyper} and \ref{sec:appendix-datasets}).

\section{Experiments}
\label{sec:Experiments}



\subsection{Opinion Target Extraction}
\label{sec:OTE}

Opinion Target Extraction (OTE) results are reported in Table \ref{char:OTE_f1score}. The zero-shot model transfer using mBERT obtains better results for Spanish and French. However, for Dutch, Russian and Turkish the best results are obtained by the data-transfer approaches. 
The overall picture changes when using XLM-RoBERTa (XLM-R) base. First, the zero-shot baseline is much closer to the gold upper bound than that of mBERT. This shows that XLM-R has better multilingual transfer learning capabilities for this task. In fact, the zero-shot transfer outperforms the translate-train and translate-test approaches for all languages except Turkish. Second, the XLM-R base results on gold-labelled data are substantially better than those of mBERT. Finally, XLM-R large offers the best cross-lingual transfer capabilities, as the zero-shot transfer is clearly superior for every language, including Turkish. 

A common trait for all three models in the OTE benchmark is that the translate-train approach is superior to the translate-test approach in the large majority of the cases. As expected, all the approaches achieve a performance significantly lower than the gold upper bound.

\subsection{Named Entity Recognition}

If we compare the OTE results with those obtained for NER (Table \ref{char:NER_f1score}), we see a number of different patterns. First, the zero-shot approach using mBERT outperforms the data-based cross-lingual transfer methods (translate-train and translate-test) for the majority of languages . Second, unlike in OTE, the translate-test is systematically better than translate-train. Third, the mBERT performance on CoNLL data is similar to that of XLM-R base. Finally, fine-tuning XLM-R base on translated and projected data obtains better results for German and Italian than the zero-shot method. However, XLM-R large provides obtains the same results as for OTE, obtaining the best results for every language in the zero-shot setting. This validates the findings of the OTE results, namely, that the performance of the zero-shot method heavily depends on the characteristics of the multilingual language model used.

\begin{table}[tbp]
  \centering
  \small
    \adjustbox{max width=\linewidth}{\begin{tabular}{|c|c|ccc|}
\hline
\multicolumn{5}{|c|}{mBERT} \\
\hline
& Gold & Zero-shot & Trans-Train & Trans-Test  \\
\hline
EN & 76.2 $\pm$ 0.9 & - & - & -  \\
ES & 75.2 $\pm$ 0.5 & \textbf{68.4} $\pm$ 0.6 & 67.9 $\pm$ 0.8 & 62.2 $\pm$ 1.2 \\
FR & 74.0 $\pm$ 1.1 & \textbf{62.7} $\pm$ 1.2 & 59.7 $\pm$ 1.2 & 57.6 $\pm$ 1.1 \\
NL & 69.7 $\pm$ 0.9 & 61.7 $\pm$ 0.8 & 64.3 $\pm$ 1.5 & \textbf{67.0} $\pm$ 0.8 \\
RU & 72.5 $\pm$ 0.5 & 53.8 $\pm$ 2.2 & \textbf{62.9} $\pm$ 0.6 & 59.7 $\pm$ 0.4 \\
TR & 62.0 $\pm$ 1.2 & 45.3 $\pm$ 4.0 & \textbf{45.7} $\pm$ 2.3 & 35.5 $\pm$ 2.4 \\

\hline\hline
\multicolumn{5}{|c|}{XLM-R base}\\
\hline
& Gold & Zero-shot & Trans-Train & Trans-Test  \\
\hline
EN & 84.4 $\pm$ 0.9 & - & - & -  \\
ES & 81.1 $\pm$ 0.7 & \textbf{78.2} $\pm$ 0.4 & 72.5 $\pm$ 0.7 & 62.9 $\pm$ 0.9 \\
FR & 80.2 $\pm$ 0.6 & \textbf{72.7} $\pm$ 0.3 & 64.7 $\pm$ 0.8 & 60.0 $\pm$ 0.6 \\
NL & 80.8 $\pm$ 1.7 & \textbf{75.5} $\pm$ 0.8 & 70.0 $\pm$ 1.6 & 71.0 $\pm$ 1.5 \\
RU & 81.5 $\pm$ 0.3 & \textbf{74.9} $\pm$ 0.9 & 69.5 $\pm$ 0.3 & 62.2 $\pm$ 1.6 \\
TR & 69.0 $\pm$ 1.1 & 58.1 $\pm$ 3.5 & \textbf{58.9} $\pm$ 1.8 & 36.4 $\pm$ 1.8 \\

\hline\hline
\multicolumn{5}{|c|}{XLM-R large} \\
\hline
& Gold & Zero-shot & Trans-Train & Trans-Test  \\
\hline
EN & 86.4 $\pm$ 1.1 & - & - & -  \\
ES & 83.6 $\pm$ 0.1 & \textbf{79.3} $\pm$ 0.8 & 73.7 $\pm$ 1.1 & 64.0 $\pm$ 1.4 \\
FR & 82.2 $\pm$ 0.6 & \textbf{74.6} $\pm$ 1.7 & 66.1 $\pm$ 0.6 & 60.7 $\pm$ 0.6 \\
NL & 80.4 $\pm$ 2.1 & \textbf{77.7} $\pm$ 1.9 & 74.0 $\pm$ 1.3 & 72.9 $\pm$ 1.8 \\
RU & 82.8 $\pm$ 0.4 & \textbf{76.8} $\pm$ 1.3 & 69.3 $\pm$ 2.3 & 62.2 $\pm$ 1.3 \\
TR & 72.3 $\pm$ 2.4 & \textbf{62.4} $\pm$ 1.0 & 57.8 $\pm$ 2.4 & 33.7 $\pm$ 0.9 \\
\hline
\end{tabular}}
  \caption{OTE F1 score with models of different capacity.}
  \label{char:OTE_f1score}
\end{table}

\begin{table}[t]
  \centering
  \small
    \adjustbox{max width=\linewidth}{\begin{tabular}{|c|c|ccc|}
\hline
\multicolumn{5}{|c|}{mBERT}\\
\hline
& Gold & Zero-shot & Trans-Train & Trans-Test  \\

\hline
EN & 90.7 $\pm$ 0.3 & - & - & - \\
ES & 87.4 $\pm$ 0.4 & \textbf{74.6} $\pm$ 0.4 & 69.5 $\pm$ 0.4 & 70.8 $\pm$ 0.6 \\
DE & 82.0 $\pm$ 0.4 & \textbf{71.0} $\pm$ 0.9 & 70.1 $\pm$ 0.3 & 70.6 $\pm$ 0.5 \\
NL & 90.8 $\pm$ 0.4 & \textbf{78.5} $\pm$ 0.5 & 74.4 $\pm$ 0.6 & 75.4 $\pm$ 0.8 \\
IT & 84.7 $\pm$ 0.3 & 68.2 $\pm$ 0.5 & 68.7 $\pm$ 0.5 & \textbf{70.7} $\pm$ 0.3 \\

\hline\hline
\multicolumn{5}{|c|}{XLM-R base}\\
\hline
& Gold & Zero-shot & Trans-Train & Trans-Test  \\
\hline
EN & 90.4 $\pm$ 0.2 & - & - & -  \\
ES & 87.7 $\pm$ 0.2 & \textbf{75.0} $\pm$ 0.4 & 70.1 $\pm$ 0.6 & 72.5 $\pm$ 0.2 \\
DE & 83.1 $\pm$ 0.3 & 67.9 $\pm$ 0.5 & \textbf{70.5} $\pm$ 0.5 & 70.1 $\pm$ 0.8 \\
NL & 89.8 $\pm$ 0.2 & \textbf{78.1} $\pm$ 0.6 & 73.3 $\pm$ 0.9 & 74.7 $\pm$ 0.4 \\
IT & 84.3 $\pm$ 0.3 & 71.2 $\pm$ 0.5 & 71.1 $\pm$ 0.4 & \textbf{71.7} $\pm$ 0.3 \\
\hline

\hline\hline
\multicolumn{5}{|c|}{XLM-R large}\\
\hline
& Gold & Zero-shot & Trans-Train & Trans-Test  \\
\hline
EN & 92.4 $\pm$ 0.1 & - & - & -  \\
ES & 88.9 $\pm$ 0.2 & \textbf{79.5} $\pm$ 1.0 & 70.9 $\pm$ 0.6 & 74.0 $\pm$ 0.5 \\
DE & 85.1 $\pm$ 0.6 & \textbf{74.5} $\pm$ 0.7 & 73.7 $\pm$ 0.5 & 72.9 $\pm$ 0.3 \\
NL & 92.9 $\pm$ 0.7 & \textbf{82.3} $\pm$ 0.6 & 77.5 $\pm$ 0.9 & 77.2 $\pm$ 0.6 \\
IT & 87.5 $\pm$ 0.2 & \textbf{76.0} $\pm$ 0.5 & 73.7 $\pm$ 0.4 & 73.5 $\pm$ 0.6 \\
\hline
\end{tabular}}
  \caption{NER F1 score with models of different capacity.}
  \label{char:NER_f1score}
\end{table}

Previous research has demonstrated that cross-lingual transfer with mBERT works best for topologically similar languages \cite{pires-etal-2019-multilingual,wu-dredze-2020-languages}, which is somewhat coherent with the results obtained for Spanish and French, where the zero-shot transfer is superior to the Translate-train and Translate-test approaches, while it is worse for Russian and Turkish. Additionally, it is worth noting that mBERT has been trained using only Wikipedia text for 104 languages.

In contrast, XLM-R (both base and large) have been trained using CommonCrawl \cite{DBLP:journals/corr/abs-1911-00359}, a much larger multilingual corpus with a variety of texts extracted from the Web, perhaps also including texts of similar domain to those in the OTE datasets. This may also account for the large differences in OTE performance between XLM-R base and mBERT. In this sense, the similar performance between mBERT and XLM-R base for NER might be partially due to the fact that the CoNLL and Evalita datasets consist of news stories in which most of the labelled entities may appear in the Wikipedia, the texts used to pre-train mBERT. 

The performance of the XLM-R large shows that pretrained models with larger capacity help to obtain strong performance across languages, also for zero-shot cross-lingual methods. Still, data-based cross-lingual transfer (Translate-Train and Translate-Test) approaches remain useful if access to the required hardware for working with such larger language models is not available. 


Finally, Table \ref{tab:zerovsSota} lists the results of previous methods that leverage parallel data and/or annotation projections to perform cross-lingual transfer on the NER CoNLL 2002-2003 data. By comparing previous work with our zero-shot baselines using  mBERT, XLM-R base and XLM-R large, we can see that the XLM-R large in the zero-shot setting still outperforms most previous approaches. The only exception being the results obtained by \citet{Li2021crosslingualNE} for German.

\begin{table}[tbp]
    \centering
    \small
    \adjustbox{max width=\linewidth}{
\begin{tabular}{l|ccc}
Models & ES & DE & NL \\
\hline

mBERT \cite{DBLP:journals/corr/abs-2101-08231} & 64.3 & - & - \\
BiLSTM + CRF \cite{jain-etal-2019-entity} & 73.5 & 61.5 & 69.9 \\
BiLSTM + CRF \cite{guo-roth-2021-constrained} & 77.9 & 71.4 & 80.6 \\
XLM-R large \cite{Li2021crosslingualNE} & 78.9 & \textbf{76.9} & 79.7 \\
\hline
mBERT (Ours - zero-shot) & 74.6 & 71.0 & 78.5 \\
XLM-R base (Ours - zero-shot) & 75.0 & 67.9 & 78.1 \\
XLM-R large (Ours - zero-shot) & \textbf{79.5} & 74.5 & \textbf{82.3} \\
\hline
XLM-R base (Ours - Translate train)  & 70.1 & 70.5 & 73.3 \\
XLM-R base (Ours - Translate test)  & 72.5 & 70.1 & 74.7 \\
XLM-R large (Ours - Translate train)  & 70.9 & 73.7 & 77.5 \\
XLM-R large (Ours - Translate test)  & 74.0 & 72.9 & 77.2 \\

\end{tabular}}
    \caption{Comparison between the previous research methods that leverage projections, the zero-shot baselines and our annotation projections in the 2002-2003 NER CoNLL datasets. F1 score reported}
    \label{tab:zerovsSota}
\end{table}

\section{Error Analysis}
\label{sec:ErrorAnalisis}

The experiments described in Section \ref{sec:Experiments} showed that 
translate-train and translate-test perform worse than the zero-shot approach when using XLM-R large. 
In this section we will assess the performance of the machine translation and word alignment models. Furthermore, we will undertake an error analysis to better understand the shortcomings of translate-train and translate-test with respect to the zero-shot cross-lingual transfer.

\subsection{Evaluating the Projection Method}\label{sec:ProjectionPerformance}

We start our experiments by analyzing the quality of our automatically projected annotations. In order to do that, human annotators manually projected the labels from the English OTE gold-labelled data to the automatic translations to Spanish, French and Russian using DeepL and M2M100 for Turkish. The annotators are NLP PhD candidates with either native and/or proficient skills in both English and the target language. See Section \ref{apen:ManualProjectionDescription} for more details. 

We compare the projections of the annotations automatically generated by the different word alignment methods with those provided by the human annotators. Table \ref{char:alignments} shows that the language model-based methods (SimAlign and AWESoME) outperform the statistically based methods (GIZA++ and FastAlign) by a wide margin in all languages. Furthermore, AWESoME consistently outperforms SimAlign for every language. 
The performance of the AWESoME system confirms that it is possible to generate high quality annotations close to those generated by human experts.
The results also show that for Turkish performance is lower than for the other languages. This is the case for the methods that require parallel data (GIZA++, FastAlign and AWESoME) as well as SimALign that does not require parallel data. So we can attribute the lower performance to the difficulty of projecting annotations for the English-Turkish pair and not the usage of the CCAligned corpus. 

\begin{table}[htbp]
  \centering
  \small
    \adjustbox{max width=\linewidth}{\begin{tabular}{c|cccc}
& GIZA++ & FastAlign & SimAlign & AWESoME \\
\hline
ES & 77.0 & 75.0 & 86.7 & \textbf{91.5} \\
FR & 73.3 & 72.9 & 86.3 & \textbf{91.3} \\
RU & 72.4 & 76.9 & 87.7 & \textbf{93.4} \\
TR & 64.0 & 68.4 & 81.9 & \textbf{88.5} \\
\end{tabular}}
  \caption{OTE F1 score between the human annotation projections vs the automatic projections generated using different alignment models.}
  \label{char:alignments}
\end{table}

While Table \ref{char:alignments} shows that we generate high quality annotation projections, the best model, AWESoME, still makes some mistakes. To explore the effect of these mistakes we fine-tune XLM-R large models on the manually projected train datasets and compare their performance on the gold-labelled test sets with the models trained on the AWESoME automatically projected data.
Table \ref{char:ManualvsAutomaticProjection} shows that the models obtained using the manually projected data are sightly better, except for Turkish, which once again acts as outlier. 
In any case, as the results obtained by fine-tuning on the manually projected data are still worse than the zero-shot method, this experiment proves that the projection of annotations is not responsible for the data-based cross-lingual transfer methods to be inferior to the zero-shot baseline.

\begin{table}[htbp]
  \centering
  \small
\begin{tabular}{c|cc}
XLMR & Trans-Train & Trans-Train (Manual) \\
\hline
ES & 73.7 $\pm$ 1.1 & \textbf{75.1} $\pm$ 1.2 \\
FR & 66.1 $\pm$ 0.6 & \textbf{67.9} $\pm$ 1.0 \\
RU & 69.3 $\pm$ 1.3 & \textbf{69.4} $\pm$ 2.1 \\
TR & \textbf{57.8} $\pm$ 2.4 & 50.6 $\pm$ 1.4 \\
\end{tabular}
  \caption{XLM-R large OTE F1 score when training with automatically and manually projected datasets}
  \label{char:ManualvsAutomaticProjection}
\end{table}

\subsection{Downstream Evaluation of Machine Translation Models}

In order to evaluate the influence of the machine translation system used, we translate the English gold-labelled data using four different translation systems. We fixed AWESoME as the word aligner for annotation projection. We fine-tune XLM-R large with each of the generated training data and evaluate it against the gold-labelled test data from OTE. As Table \ref{char:translators} shows, there are no big differences in the final F1 scores when using different translation systems (Turkish is again being the exception), we decided to carry on using DeepL based on the manual assessment mentioned in Section \ref{sec:Method}.  

\begin{table}[htbp]
  \centering
  \small
    \adjustbox{max width=\linewidth}{\begin{tabular}{c|cccc}
& MarianMT & Mbart & M2M100 & DeepL \\
\hline
ES & \textbf{75.6} $\pm$ 0.8 & 75.3 $\pm$ 0.7 & 74.2 $\pm$ 0.8 & 73.7 $\pm$ 1.1 \\
FR & 64.5 $\pm$ 1.6 & \textbf{66.4} $\pm$ 1.1 & 64.9 $\pm$ 1.3 & 66.1 $\pm$ 0.6 \\
NL & 70.0 $\pm$ 2.0 & 68.8 $\pm$ 4.0 & 70.1 $\pm$ 3.1 & \textbf{74.0} $\pm$ 1.3 \\
RU & 66.6 $\pm$ 4.4 & \textbf{69.7} $\pm$ 1.4 & \textbf{69.7} $\pm$ 0.7 & 69.3 $\pm$ 2.3 \\
TR & 49.5 $\pm$ 2.9 & 56.1 $\pm$ 5.2 & \textbf{57.8} $\pm$ 2.4 & - \\
\end{tabular}}
  \caption{OTE F1 score of different XLM-R large models trained using data generated with different translation systems.}
  \label{char:translators}
\end{table}

\subsection{Where do the models fail?}

To better understand what is happening we identify the most common false negatives and positives for both OTE and NER tasks. Table \ref{tab:FalseNegatives} shows the most frequent false negatives and positives where there is a big mismatch between methods.

As it has been previously noticed \cite{AGERRI201985}, in the ABSA data the words ``comida'' (food) and ``restaurante'' (restaurant) are highly ambiguous, so we could expect the models to fail with these words. In addition, we have found out 4 main sources of errors, which are analyzed below.

\paragraph{Many-to-one translations:} This is stereotypical of targets such as ``trato'' and ``atención'' in Spanish, which, in addition to ``servicio'', are used to refer to ``service'' in English. There are 160 sentences in the English gold-labelled data containing the word ``service''; in 153 of them ``service'' is labelled as target. DeepL systematically translates it as ``servicio''. However, as shown by Table \ref{tab:WordNoDataset}, in the Spanish gold-labelled data ``service'' is also commonly referred as ``trato'' or ``atención'', instead of ``servicio''.

This would result in a training set without any occurrences of ``trato'' and ``atención'' which often occur in the gold-labelled test data. Both the zero-shot and the data-based cross-lingual transfer approaches fail to correctly label these words, which shows a problem of using automatically translated data. Interestingly, the zero-shot approach using XLM-R large correctly classifies ``trato'' (only fails to label 1 of the 19 occurrences). As shown by our experimental results, XLM-R large is more robust than mBERT and XLM-R base.

Something similar happens with the word ``place'', which in Spanish can be most frequently translated as ``lugar''  or ``sitio''. However, DeepL almost always translates it as ``lugar'' which results in ``sitio'' being absent in the automatically generated training data while being more frequent than ``lugar'' in the gold-labelled data. Note that this is not a problem for the ``translate-test'', given that the translation direction is Spanish to English.

\paragraph{Errors induced by incorrect or missing alignments:} For NER we found errors of different nature. Articles and prepositions (i.e. ``de'', ``la'') are among the words with higher false positive rate for the translate-test and translate-train approaches. We can attribute it to word alignment errors. Large multi-word named entities such as ``Consejo General de la Arquitectura Técnica de España'' (General Council of Technical Architecture of Spain) are labelled as entities. Word aligners struggle to correctly align articles in these complex expressions specially when a one-to-many or many-to-one alignment is required. In fact, in this example, the word aligners we tested failed to correctly align ``of'' with ``de la''.

\paragraph{Errors induced by dataset inconsistencies:} Another issue is the differences across languages in the original gold-labelled annotations. Thus, ``Gobierno'' (Government) and ``Estado'' (State) are labelled as organizations in the Spanish gold-labelled data, but they are not considered to be entities in the English gold-labelled data. The opposite occurs with demonym words. They are labelled as miscellaneous entities in the English data but in Spanish they are not annotated. Cross-lingual models are likely to fail labelling these cases.

\paragraph{Lost in Translation:} Finally, there is another group of words related to Spanish Government names which are not commonly used in English for the same contexts (i.e. ``Economía'' to refer to the ``Ministry of Economy'' or ``Ministerio de Economía'' in Spanish, ``Junta'' for ``local government'', or ``Plan'' for ``government projects''). While these words appear frequently in the Spanish data as part of commonly used named entities, that is not the case in the English data, where it is customary to use ``Treasury Department'' (or variations thereof) which are correctly translated into Spanish by DeepL as ``Departamento del Tesoro''. This means that, during fine-tuning on the translated data, the model is not receiving any signal to learn that ``Economy'' may be part of an organization entity. This may explain why the zero-shot method performs better for cases such as ``Economía'', ``Hacienda'', ``Plan'' and ``Junta'', listed in Table \ref{tab:FalseNegatives}.

\newcolumntype{R}[1]{>{\raggedright\arraybackslash}p{#1}}
\begin{table*}[tbp]
    \centering
    \small
\adjustbox{max width=\linewidth}{
\begin{tabular}{l|R{0.15cm}R{0.15cm}R{0.15cm}|R{0.15cm}R{0.15cm}R{0.15cm}|R{0.32cm}R{0.32cm}R{0.32cm}|R{0.32cm}R{0.32cm}R{0.32cm}|R{0.5cm}}
& \multicolumn{3}{c}{GOLD} & \multicolumn{3}{c}{Zero-shot} & \multicolumn{3}{c}{Tr-Train} & \multicolumn{3}{c}{Tr-Test} & Total \\
& B & Xb & Xl & B & Xb & Xl & B & Xb & Xl & B & Xb & Xl & \\
\hline
\multicolumn{13}{c}{OTE False Negatives } & \\
\hline
comida & 3 & 3 & 2 & 6 & 2 & 1 & 4 & 1 & 1 & 1 & 9 & 5 & 98 \\
restaurante & 7 & 5 & 7 & 9 & 5 & 6 & 7 & 6 & 6 & 7 & 12 & 10 & 43 \\
servicio & 2 & 2 & 2 & 2 & 1 & 1 & 2 & 0 & 1 & 1 & 1 & 2 & 85 \\
trato & 1 & 1 & 0 & 5 & 6 & 1 & 14 & 10 & 5 & 6 & 8 & 6 & 19 \\
atención & 2 & 3 & 3 & 8 & 2 & 3 & 7 & 1 & 3 & 7 & 7 & 7 & 13 \\
lugar & 0 & 0 & 0 & 2 & 0 & 0 & 1 & 0 & 0 & 0 & 1 & 0 & 12 \\
sitio & 1 & 0 & 0 & 5 & 1 & 1 & 3 & 3 & 3 & 2 & 1 & 1 & 14 \\
\hline
\multicolumn{13}{c}{NER False Negatives} & \\
\hline
de & 32 & 29 & 33 & 45 & 51 & 90 & 233 & 252 & 264 & 148 & 146 & 167 & 450 \\
la & 4 & 5 & 3 & 10 & 12 & 16 & 63 & 62 & 62 & 45 & 44 & 45 & 174 \\
Gobierno & 0 & 0 & 0 & 17 & 53 & 64 & 72 & 70 & 75 & 30 & 45 & 67 & 80 \\
Estado & 0 & 0 & 0 & 4 & 4 & 8 & 9 & 8 & 9 & 6 & 6 & 8 & 10 \\
Administación & 0 & 0 & 0 & 4 & 8 & 11 & 10 & 11 & 11 & 5 & 5 & 7 & 11 \\
Economía & 0 & 0 & 0 & 2 & 6 & 2 & 7 & 8 & 8 & 5 & 6 & 8 & 8 \\
Plan & 0 & 0 & 0 & 1 & 2 & 2 & 3 & 5 & 5 & 1 & 4 & 7 & 8 \\
Junta & 0 & 0 & 0 & 0 & 0 & 0 & 4 & 10 & 8 & 2 & 3 & 5 & 24 \\
Hacienda & 0 & 0 & 0 & 1 & 3 & 0 & 4 & 4 & 4 & 4 & 3 & 4 & 5 \\
\hline
\multicolumn{13}{c}{NER False Positives} & \\
\hline
español & 0 & 0 & 0 & 16 & 16 & 2 & 16 & 16 & 12 & 13 & 14 & 15 & 0 \\
catalán & 0 & 0 & 0 & 8 & 8 & 5 & 7 & 7 & 8 & 8 & 8 & 8 & 0 \\
\end{tabular}

}
    \caption{Most common false negatives and positives were there is a big mismatch between methods and the total number of labelled apperances of the word in the test data. B is the acronym for mBERT, Xb for XLM-R base and Xl for XLM-R large.}
    \label{tab:FalseNegatives}
\end{table*}

\begin{table}[tbp]
    \centering
    \small
\adjustbox{max width=\linewidth}{\begin{tabular}{cc|ccc}
En.Word & Es.Word & En Gold & Es Gold & Es Translate \\
\hline
Service & Servicio & 153 & 229 & 133 \\
Treatment & Trato & 0 & 54 & 0 \\
Attention & Atención & 2 & 35 & 0 \\
\hline
Place & Sitio & 120 & 41 & 2 \\
Place & Lugar & 120 & 19 & 91 \\
\end{tabular}}

    \caption{Number of times words appear as target words in the train datasets}
    \label{tab:WordNoDataset}
\end{table}




Summarizing, we see that machine translation data often generates a signal which is, due to inherent differences in language use, different to the signal received when using gold-labelled data in the target language. This disagreement seems to be the most common reason for the larger number of false positive and negatives of the data-based cross-lingual transfer method with respect to the zero-shot technique.


\section{Concluding Remarks}
\label{sec:Conclusions}

In this paper we described an in-depth and comprehensive evaluation of model-based and data-based zero-resource cross-lingual sequence labelling on two different tasks. 

Contrary to what previous research suggests, zero-shot transfer approach is the best performing method when using high capacity multilingual language models such as XLM-R large. However, data-based cross-lingual transfer approaches are still useful when having a model with poor downstream cross-lingual performance. For example, when using a pretrained language model not trained for a specific domain, or when the required hardware for working with such larger language models is not readily available. 


A detailed error analysis demonstrates that data-based cross-lingual transfer is hindered by machine translations which, although linguistically sound, do not align with the cultural behaviour of the target language use. Moreover, the results also show that the different word alignments methods (for annotation projection) are of high quality, obtaining comparable results with respect to manually generated alignments.



In any case, our results establish that there is still room for improving the cross-lingual performance of zero-resource sequence labelling. 





\section*{Acknowledgments}

We are grateful to Nayla Escribano, Suna Şeyma Uçar and Olia Toporkov for their help in manually projecting the labels from the English gold-labelled data into the automatically translated dataset.
We are also thankful to the anonymous reviewers for their insightful comments. This work has been supported by the following projects: (i) DeepKnowledge (PID2021-127777OB-C21) funded by MCIN/AEI/10.13039/501100011033 and FEDER Una manera de hacer Europa; (ii) Disargue (TED2021-130810B-C21), MCIN/AEI/10.13039/501100011033 and European Union NextGenerationEU/PRTR; (iii) DeepR3 (TED2021-130295B-C31) funded by MCIN/AEI/10.13039/501100011033 and EU NextGeneration programme EU/PRTR; (iv) Antidote (PCI2020-120717-2) funded by MCIN/AEI/10.13039/501100011033 and by European Union NextGenerationEU/PRTR. Iker García-Ferrero is supported by a doctoral grant from the Basque Government (PRE\_2021\_2\_0219). Rodrigo Agerri's work is also supported by the RYC-2017-23647 fellowship (MCIN/AEI/10.13039/501100011033 y por El FSE invierte en tu futuro).

\section*{Limitations}
We compare baseline cross-lingual zero-shot model transfer with machine translation and annotation projection. We do not explore alternative cross-lingual data-based methods, such as the usage of available parallel corpora instead of a machine translated corpus. We also skip evaluating methods to improve model-transfer approaches such as the ones described in Section \ref{subsec:model_transfer}. 
We may also consider that our annotation projection approach and zero-shot model transfer approach work for Indo-European languages, while their performance for other language families remains unknown. Finally, the error analysis was performed for the EN-ES language pair only.

In any case, we believe that our main claim still holds. Even though MT quality has substantially improved over the last few years, our results indicate the current optimal solution to perform cross-lingual transfer is by using large multilingual language models such as XLM-RoBERTa-large. Thus, our error analysis suggests that this might be due to important differences in language use. More specifically, MT often generates a textual signal which is different to what the models are exposed to when using gold standard data, which affects both the fine-tuning and evaluation processes. This is confirmed by our error analysis which shows that mistranslations are not the main source of errors in the data-transfer method.

\bibliography{custom}
\bibliographystyle{acl_natbib}

\appendix
\label{apen:Apendix}

\section{Model size}\label{apen:ModelSize}
We experiment with multilingual BERT (mBERT) \cite{DBLP:conf/naacl/DevlinCLT19} and XLM-RoBERTa (XML-R) base and large \cite{xlmr}. We list the number of parameters of each model in Table \ref{tab:ParamNo}
\begin{table}[htbp]
    \centering
\begin{tabular}{l|c}
Model & \#params \\
\hline 
multilingual BERT & 110M \\
XLM-RoBERTa-base & 250M \\
XLM-RoBERTa-large & 560M \\
\end{tabular}
    \caption{Number of parameters for the language models that we use in our experiments}
    \label{tab:ParamNo}
\end{table}

\section{Hyper parameters}\label{sec:appendix-hyper}

\subsection{Word alignment models}
We train AWESoME with $8$ batch size and $2e-5$ learning rate for $40,000$ steps, with all the unsupervised training objectives (mlm,tlm,tlm\_full,so,psi) and softmax extraction method. We use mBERT as backbone. 
For SimAlign we run inference with 0.0 distortion rate, 1.0 null align rate and the "itermax" matching method. We use bpe tokens and mBERT backbone. 
We use the MGIZA multicore implementation \footnote{\url{https://github.com/moses-smt/mgiza}} of GIZA++ with the recommended configuration file \footnote{\url{https://pastebin.com/b1ksHtUy}}. We use FastAlign with the default hyper-parameters. For both, GIZA++ and FastAlign we combine the forward and backward directions of the alignments using the grow-diag-final-and algorithm.

\subsection{Sequence Labelling models}\label{sec:appendix-models}
For OTE we use a batch size of $32$, $5e-5$ learning rate, we train the model for $10$ epochs and $128$ maximum sequence length. Since only a train and test splits are available for the OTE task, we use the train set as both, train and development data. 
For NER we use a batch size of $32$, $2e-5$ learning rate, we train the model for $4$ epochs and 256 maximum sequence length. We use the default values (sequence labelling implementation of the Huggingface library \footnote{\url{https://github.com/huggingface/transformers/tree/main/examples/pytorch/token-classification}}) for the remaining hyperparameters. 
For both tasks we use the BILOU encoding scheme. 

\section{Datasets Size}\label{sec:appendix-datasets}

We list the dataset size (number of sentences) of the datasets we use. 

For OTE we use the SemEval-2016 Task 5 Aspect Based Sentiment Analysis (ABSA) datasets \cite{pontiki-etal-2016-semeval}. We list the size of the datasets in Table \ref{tab:OteDatasetSize}.

\begin{table}[htbp]
    \centering
\begin{tabular}{c|ll}
Lang & Train & Test \\
\hline
EN & 2000 & 676 \\
ES & 2070 & 881 \\
FR & 1664 & 668 \\
NL & 1722 & 575 \\
RU & 3655 & 1209 \\
TR & 1232 & 144 \\
\end{tabular}
    \caption{Number of sentences in the OTE datasets}
    \label{tab:OteDatasetSize}
\end{table}

For NER we use the Spanish and Dutch data from the CoNLL-2002 datasets \cite{tjong-kim-sang-2002-introduction}. For English and German we use the CoNLL-2003 datasets \cite{tjong-kim-sang-de-meulder-2003-introduction} and for Italian we use Evalita 2009 data \cite{speranza2009named}. We list the size of these datasets in Table \ref{tab:NerDatasetSize}.

\begin{table}[htbp]
    \centering
\begin{tabular}{c|lll}
& Train & Dev & Test \\
\hline
EN & 14987 & 3466 & 3684 \\
ES & 6871 & 1914 & 1516 \\
DE & 12705 & 3068 & 3160 \\
NL & 15806 & 2895 & 5195 \\
IT & 11227 & 0 & 4136 \\
\end{tabular}
    \caption{Number of sentences in the NER datasets}
    \label{tab:NerDatasetSize}
\end{table}

\section{Computer infrastructure}
We perform all our experiments using a single NVIDIA A30 GPU with 24GB memory. The machine used has two Xeon Gold 6226R CPUs and 256GB RAM. 

\section{Manual Projection of the datasets}
\label{apen:ManualProjectionDescription}
\begin{figure*}[t]
    \centering
    \includegraphics[width=0.80\linewidth]{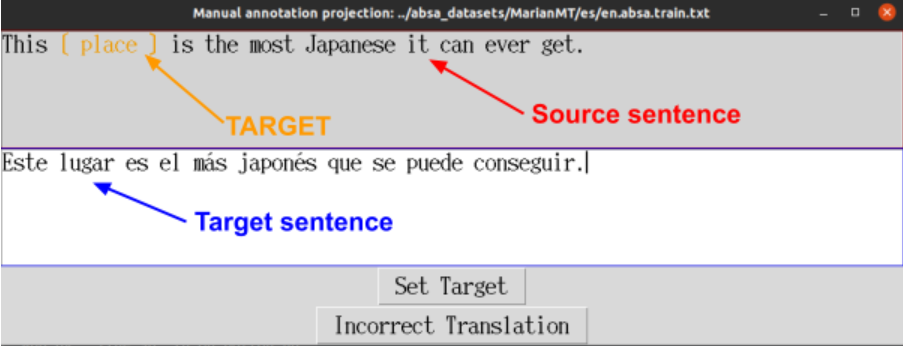}
    \includegraphics[width=0.80\linewidth]{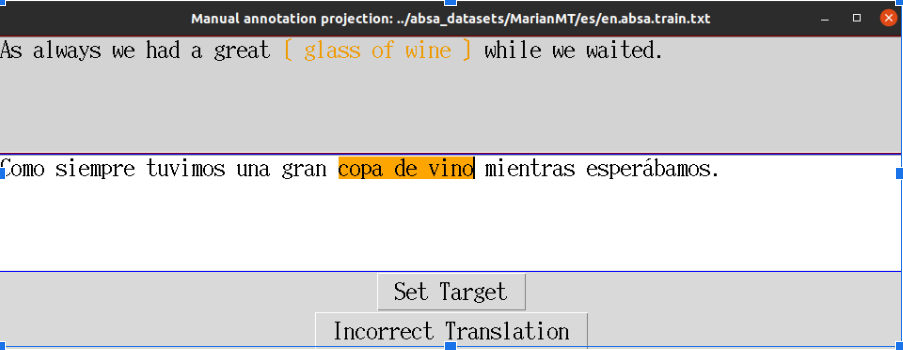}
    \caption{Application used to manually annotate the projections}
    \label{fig:AppImages}
\end{figure*}

Human annotators manually projected the labels from the English OTE gold data to the automatic translations to Spanish, French and Russian using DeepL and m2m10 for Turkish
The annotators are NLP PhD candidates with either native and/or proficient skills in both English and the target language. We describe the experiment in Section \ref{sec:ProjectionPerformance}.
For the purpose of this experiment, we developed an application to assist during the annotation process. The annotator sees the sentence in English, where there is a highlighted target and must select the same target in a translated target sentence. Figure \ref{fig:AppImages} shows two screenshots from the application. The full guidelines and the code of the application provided to the annotators are available at \url{https://github.com/ikergarcia1996/Annotation-Projection-App}. 

\end{document}